\documentclass{article}
\usepackage{tikz, xcolor}
\usepackage{spconf,amsmath,graphicx}
\usepackage{booktabs}       
\usepackage{multirow}       
\usepackage{graphicx}       
\usepackage{threeparttable} 
\usepackage{array}
\usepackage{amsmath}
\usepackage{amsfonts}
\usepackage{tabularx}
\usepackage{xspace}
\usepackage{orcidlink}
\usepackage{makecell}
\usepackage{caption}
\usepackage{stfloats}
\usepackage{cuted}
\usepackage{enumitem}
\usepackage{fontawesome5}

\definecolor{gold}{RGB}{255, 215, 0}
\definecolor{silver}{RGB}{192, 192, 192}
\definecolor{bronze}{RGB}{205, 127, 50}

\newcommand{\rankbadge}[2]{%
  \tikz[baseline=(n.base)]\node[
    fill=#1!25!white,
    draw=#1!40!black,
    text=black,
    rounded corners=2pt,
    inner sep=1pt,
    minimum width=0.8em
  ](n){#2};%
}

\newcommand{\up}[1]{\textcolor{teal!70!black}{\scriptsize$\uparrow$\,#1}}

\newcommand{\methname}{AVA-DINO\xspace}

\title{Anomaly-Aware Vision-Language Adapters for Zero-Shot \\Anomaly Detection}
\name{Muhammad Aqeel$^{1, *}$\orcidlink{0009-0000-5095-605X}, Maham Nazir$^{2, *}$\orcidlink{0009-0004-1832-297X}, Uzair Khan$^{1}$\orcidlink{0000-0003-4107-4359}, Marco Cristani$^{1,3}$\orcidlink{0000-0002-0523-6042}, Francesco Setti$^{1}$\orcidlink{0000-0002-0015-5534}}

\address{$^{1}$ Dept. of Engineering for Innovation Medicine, University of Verona, Italy \\
$^{2}$ School of Computer Science and Engineering, Beihang University, China \\
$^{3}$ Dept. of Computer Science, Reykjavik University, Iceland \\
$^{*}$ Equal Contribution \\
muhammad.aqeel@univr.it}
\begin{document}
%
\maketitle
\begin{abstract}
Zero-shot anomaly detection aims to identify defects in unseen categories without target-specific training. Existing methods usually apply the same feature transformation to all samples, treating normal and anomalous data uniformly despite their fundamentally asymmetric distributions, compact normals versus diverse anomalies. We instead exploit this natural asymmetry by proposing \methname, an anomaly-aware vision-language adaptation framework with dual specialized branches for normal and anomalous patterns that adapt frozen DINOv3 visual features. During training on auxiliary data, the two branches are learned jointly with a text-guided routing mechanism and explicit routing regularization that encourages branch specialization. At test time, only the input image and fixed, predefined language descriptions are used to dynamically combine the two branches, enabling an asymmetric activation. This design prevents degenerate uniform routing and allows context-specific feature transformations. Experiments across nine industrial and medical benchmarks demonstrate state-of-the-art performance, achieving 93.5\% image-AUROC on MVTec-AD and strong cross-domain generalization to medical imaging without domain-specific fine-tuning. 
\href{https://github.com/aqeeelmirza/AVA-DINO}{\faGithub~AVA-DINO Github Repo}
\end{abstract}
\section{Introduction}
\label{sec:intro}

Industrial quality inspection demands detecting diverse defect types across varying object categories, yet collecting exhaustive anomaly samples for supervised training remains impractical due to the rarity and diversity of failure modes. Zero-shot anomaly detection (ZSAD) addresses this challenge by leveraging vision-language models pre-trained on large-scale datasets, where language embeddings provide semantic priors that enable generalization to unseen categories without target-specific training data~\cite{jeong2023winclip,zhou2024anomalyclip}. 

\begin{figure}[t]
\centering
\includegraphics[width=\columnwidth]{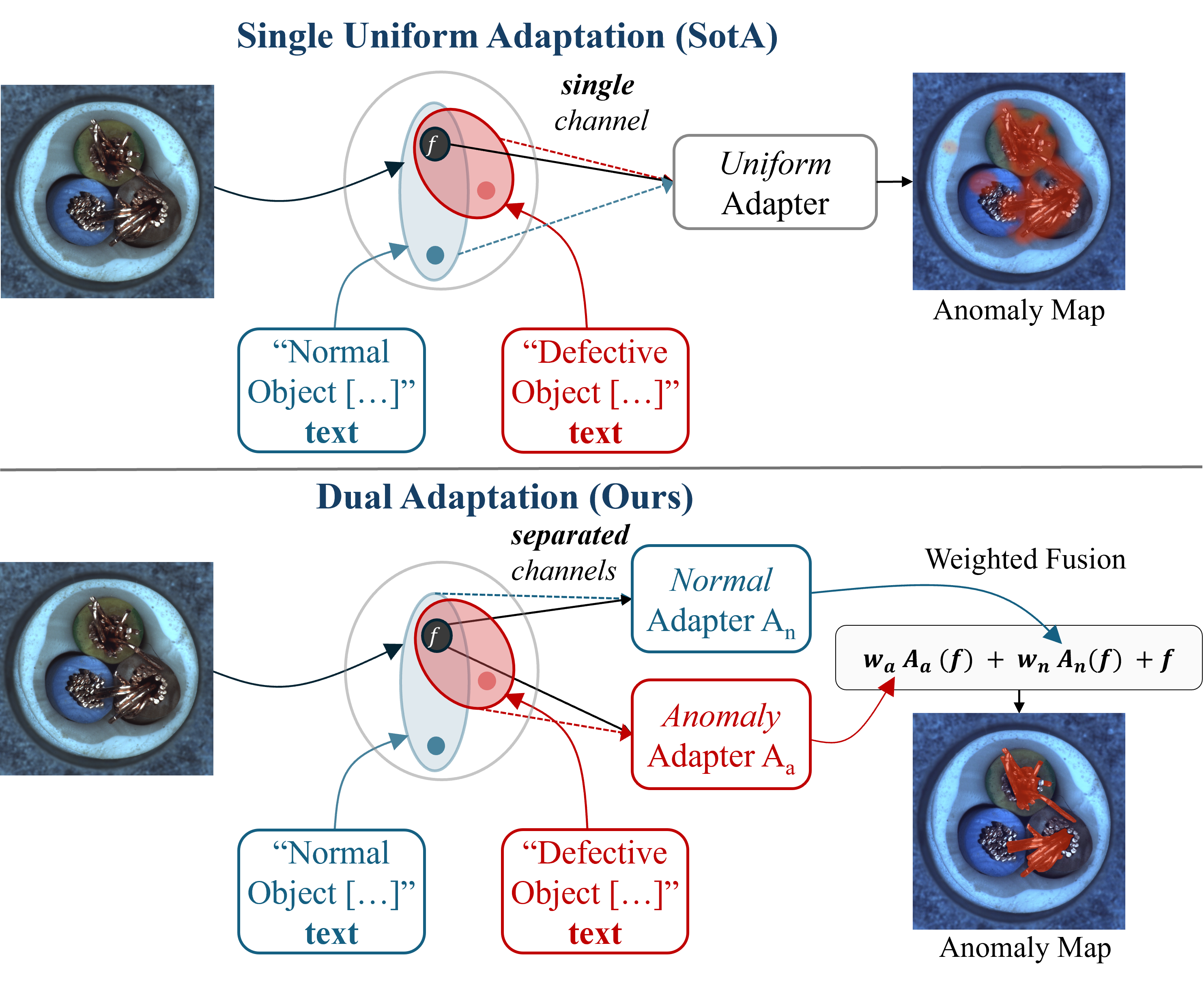}
\caption{Uniform adaptation transforms all samples identically through a single adapter, producing mixed features (top). Our dual adaptation routes feed embeddings \textbf{f} into specialized normal and anomaly adapters, providing 1) separated representations $\mathcal{A}_n(\mathbf{f})$, $\mathcal{A}_a(\mathbf{f})$ and 2) weights $w_n$, $w_a$ to be fused together (bottom). Both the frameworks are shown at test time.}
\label{fig:radar}
\end{figure}

In a typical ZSAD pipeline, a frozen vision backbone extracts global and patch-level features from test images, which are directly compared to text or visual prototypes encoding normality, yielding image-level scores and pixel-level anomaly maps without any target-domain supervision. Following standard ZSAD protocols, all supervision is confined to auxiliary datasets, while target domains require no labels or adaptation. Recent advances employ adapter networks to efficiently fine-tune frozen foundation models such as CLIP~\cite{radford2021learning} and DINOv3~\cite{simeoni2025dinov3}, enabling cross-domain anomaly detection through lightweight parameter updates. However, existing ZSAD approaches rely on single-branch adaptation and apply uniform feature transformations to all samples, failing to capture the fundamental asymmetry between normal patterns and anomalous patterns (Figure~\ref{fig:radar}).
We observe that normal and anomalous samples require fundamentally different feature transformations in the adaptation process. Normal samples benefit from transformations that emphasize subtle texture variations while preserving object structure and appearance consistency, establishing compact feature clusters that facilitate outlier detection. Conversely, anomalies require transformations that highlight deviations, amplify defect boundaries, and capture structural irregularities spanning diverse failure modes. This asymmetry motivates our core insight: rather than learning a single adaptation pathway that compromises between these conflicting objectives, we should learn separate, specialized transformations for each context and dynamically route samples to appropriate pathways based on their semantic characteristics.

We propose~\textbf{\emph{\methname}} (\emph{\textbf{A}}nomaly-Aware \emph{\textbf{V}}ision Language \emph{\textbf{A}}dapters with \emph{\textbf{DINO}}v3), a novel architecture where separate adaptation modules learn normal-specific and a\-no\-ma\-ly-spe\-ci\-fic transformations of DINOv3 features through vision-language contrastive learning. Our method employs frozen DINOv3 for visual feature extraction and frozen CLIP for semantic text encoding, introducing lightweight auxiliary-trained adapters organized into dual pathways that refine multi-scale DINOv3 representations. At test time, a text-guided routing mechanism computes semantic similarity between DINOv3 features and CLIP text embeddings (e.g., "a photo of perfect hazelnut" versus "a photo of damaged hazelnut"), generating dynamic routing weights via temperature-scaled softmax that determine each pathway's contribution to feature adaptation. To learn this mechanism, we introduce a routing regularization loss that, at training time, encourages pathway specialization by supervising routing decisions on auxiliary data, ensuring normal samples predominantly activate the normal adapter while anomalies engage the anomaly adapter. This explicit training-only supervision prevents routing collapse to uniform 50-50 weighting and enables interpretable pathway selection that reveals which adaptation context applies to each sample.

Our key contributions are as follows:
\begin{itemize} [leftmargin=*, topsep=2pt, itemsep=1pt, parsep=0pt, partopsep=0pt]
    \item We introduce an anomaly-aware dual-branch adaptation framework that learns context-specific transformations for normal and anomalous patterns, overcoming the limitation of uniform adaptation in existing ZSAD methods.
    \item We propose text-guided dynamic routing with routing regularization that ensures pathway specialization via training only auxiliary-data supervision, preventing routing collapse while enabling interpretable pathway selection.
    \item We demonstrate impressive cross-domain generalization across industrial and medical benchmarks as common in recent literature~\cite{qu2025bayesian}, with routing behavior analysis confirming learned specialization that adapts to sample characteristics.
\end{itemize}

\section{Related Work}
\label{sec:relatedwork}

Recent advances in vision-language models have enabled zero-shot anomaly detection without target-specific training data. WinCLIP~\cite{jeong2023winclip} applies CLIP features with learnable text prompts in a sliding window manner for localized anomaly detection. AnomalyCLIP~\cite{zhou2024anomalyclip} introduces object-aware prompt learning to enhance semantic alignment between visual and textual representations. AdaCLIP~\cite{cao2024adaclip} proposes hybrid semantic embeddings for improved cross-domain transfer, while Bayes-PFL~\cite{qu2025bayesian} employs Bayesian prompt fine-tuning for uncertainty-aware detection. More recently, CoZAD~\cite{aqeel2025contrastive} combines contrastive learning with confident meta-learning for improved zero-shot generalization. However, all these methods employ uniform feature adaptation during both training and testing, applying identical transformations regardless of whether samples exhibit normal or anomalous characteristics. Our work addresses this fundamental limitation by introducing context-aware dual-branch adaptation.

Adapter modules~\cite{houlsby2019parameter, chen2022adaptformer} enable efficient fine-tuning of large pre-trained models by inserting bottleneck layers while keeping backbone parameters frozen. This paradigm has proven effective across NLP and vision tasks. CLIP-Adapter and Tip-Adapter demonstrate adapter efficacy for vision-language models. However, existing adapter architectures apply uniform transformations to all inputs within a task. Our anomaly-aware adapters represent a new strategy by learning multiple specialized pathways with dynamic routing, enabling context-specific feature transformations that adapt to the fundamental asymmetry between normal and anomalous patterns.

\section{Proposed Approach}
\label{sec:method}
\begin{figure*}[t]
\centering
\includegraphics[width=\textwidth]{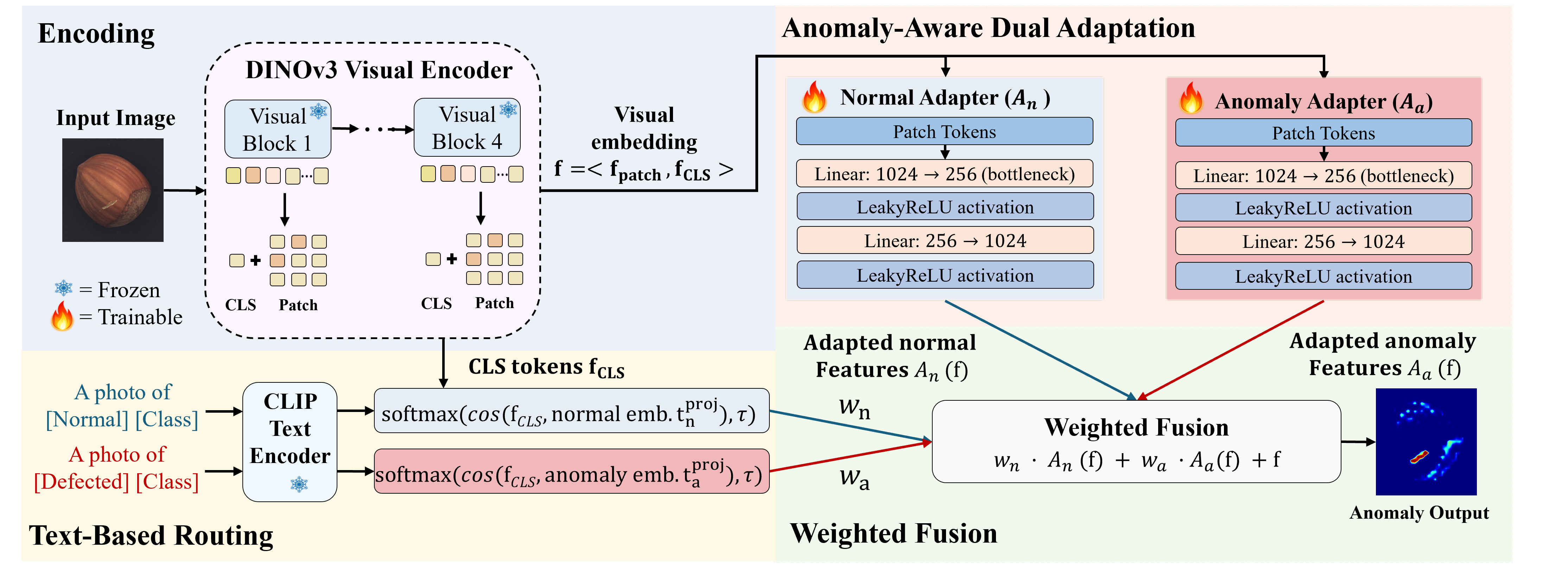}
\caption{Architecture of \methname. During training, the frozen DINOv3 visual encoder extracts multi-scale features from input images. Patch tokens are processed by dual-branch adapters (normal and anomaly pathways), while CLS tokens guide text-based routing via the frozen CLIP text encoder. Routing weights ($w_n$, $w_a$) dynamically combine adapted features to generate pixel-level anomaly maps and image-level predictions. During testing, the learned routing weights automatically steer features toward the appropriate pathway, producing anomaly maps and predictions on unseen images without further adaptation.}
\label{fig:architecture}
\end{figure*}

\subsection{Overview}
\label{Sec:Overview}
Zero-shot anomaly detection aims to generate a pixel-wise anomaly map $\mathbf{M} \in [0,1]^{H \times W}$ for a test image $\mathbf{x} \in \mathbb{R}^{H \times W \times 3}$ without exposure to target-specific samples during training. Following established protocols, we train on an auxiliary dataset $\mathcal{D}_a$ containing normal and anomalous samples with ground-truth masks, then evaluate on a disjoint target dataset $\mathcal{D}_t$ where $\mathcal{D}_a \cap \mathcal{D}_t = \emptyset$.
Fig.~\ref{fig:architecture} reports the detailed scheme of \methname, which is composed by four modules: 1) Encoding, 2) Anomaly-aware Dual Adaptation,  3) Text-based Routing, 4) Weighted Fusion. 
In the following, we present the method in its training stage. The inference is sketched in the teaser (Fig.~\ref{fig:radar}).\\

\subsection{\methname Pipeline}
\label{Sec:Pipeline}
\noindent\textbf{Encoding.} The encoding module receives the input image and encodes it with a frozen DINOv3 Vision Transformer. The image is partitioned into patches and mapped to a sequence of patch tokens, augmented with a CLS token. After passing through the stacked DINOv3 visual blocks, the encoder outputs the embedding $\mathbf{f}$ composed by the local patch features $\mathbf{f}_{\text{patch}} \in \mathbb{R}^{1024}$ and a global representation $\mathbf{f}_{\text{CLS}}$, which is the CLS token $\mathbf{f}_{\text{cls}} \in \mathbb{R}^{1024}$ from DINOv3's final layer. \\

\noindent\textbf{Anomaly-aware Dual Adaptation.} We introduce dual lightweight adapters that recalibrate representations toward anomaly-specific discrimination. The \textit{normal adapter} $\mathcal{A}_n$ learns transformations preserving structural consistency characteristic of defect-free samples, while the \textit{anomaly adapter} $\mathcal{A}_a$ learns transformations emphasizing deviations and defect boundaries. Given visual features $\mathbf{f}$ comprising patch tokens and CLS token from DINOv3, each branch produces context-specific representations:

\begin{equation}
\mathbf{f}_n = \mathcal{A}_n(\mathbf{f}), \quad \mathbf{f}_a = \mathcal{A}_a(\mathbf{f})
\end{equation}

\noindent This parallel processing enables each branch to develop distinct feature transformations optimized for its assigned context, with text-guided routing determining how to combine them.\\

\noindent\textbf{Text-Guided Dynamic Routing.}
Rather than manually selecting pathways or naively averaging outputs, we leverage CLIP's semantic understanding to selectively weight each branch. During training, text descriptions of normal and anomalous states provide semantic anchors for determining which transformation best suits each input. We exploit CLIP to obtain $\mathbf{t}_n$, which encodes a normal prompt (``a photo of [normal] [class]'', e.g., \textit{perfect hazelnut})  and $\mathbf{t}_a$, encoding an anomalous prompt  (``a photo of [defected] [class]'', e.g., \textit{damaged hazelnut}). Subsequently, we project these embeddings to align with DINOv3's feature space:
\begin{equation}
\mathbf{t}_n^{\text{proj}} = \mathbf{W}_{\text{proj}} \mathbf{t}_n, \quad \mathbf{t}_a^{\text{proj}} = \mathbf{W}_{\text{proj}} \mathbf{t}_a
\end{equation}
where $\mathbf{W}_{\text{proj}}$ is a learnable projection matrix. We extract the CLS token $\mathbf{f}_{\text{cls}} \in \mathbb{R}^{1024}$ from DINOv3's final layer as a global image representation. Routing weights are computed at inference time via temperature-scaled softmax over cosine similarities between $\mathbf{f}_{\text{cls}}$ and projected text embeddings:
\begin{equation}
[w_n, w_a] = \text{softmax}\left(\frac{[\text{cos}(\mathbf{f}_{\text{cls}}, \mathbf{t}_n^{\text{proj}}), \text{cos}(\mathbf{f}_{\text{cls}}, \mathbf{t}_a^{\text{proj}})]}{\tau}\right)
\end{equation}.\\

\noindent\textbf{Weighted Fusion.}
The final adapted features integrate both branches proportionally with a residual connection preserving original representations:
\begin{equation}
\mathbf{f}_{\text{adapted}} = w_n \cdot \mathbf{f}_n + w_a \cdot \mathbf{f}_a + \mathbf{f}
\end{equation}
where $\mathbf{f}$ denotes the original DINOv3 patch features. This enables smooth interpolation between normal and anomaly-specific transformations while maintaining gradient flow through the residual path.

\begin{table*}[t]
\centering
\caption[Zero-shot anomaly detection performance]{Zero-shot anomaly detection performance across benchmarks. Rankings: \rankbadge{gold}{1st}, \rankbadge{silver}{2nd}, \rankbadge{bronze}{3rd} for each metric per dataset.}
\label{tab:main_results}
\resizebox{\textwidth}{!}{
\begin{tabular}{ll|ccc|ccc|ccc|ccc|ccc}
\toprule
\multirow{2}{*}{\textbf{Domain}} & \multirow{2}{*}{\textbf{Dataset}} & 
\multicolumn{3}{c|}{\textbf{WinCLIP (2023)}~\cite{jeong2023winclip}} & 
\multicolumn{3}{c|}{\textbf{AnomalyCLIP (2024)}~\cite{zhou2024anomalyclip}} & 
\multicolumn{3}{c|}{\textbf{AdaCLIP (2024)}~\cite{cao2024adaclip}} & 
\multicolumn{3}{c|}{\textbf{Bayes-PFL (2025)}~\cite{qu2025bayesian}} & 
\multicolumn{3}{c}{\textbf{\methname (Ours)}}
\\
\cmidrule(lr){3-5} \cmidrule(lr){6-8} \cmidrule(lr){9-11} \cmidrule(lr){12-14} \cmidrule(lr){15-17}
& & I-AUC & P-AUC & Pixel-F1 & I-AUC & P-AUC & Pixel-F1 & I-AUC & P-AUC & Pixel-F1 & I-AUC & P-AUC & Pixel-F1 & I-AUC & P-AUC & Pixel-F1 \\
\midrule
\multirow{7}{*}{\textbf{Industrial}} 
& MVTec-AD    & 91.8 & 85.1 & 31.6 & 91.5 & \rankbadge{bronze}{90.9} & 39.1 & \rankbadge{bronze}{92.0} & 88.7 & \rankbadge{bronze}{43.4} & \rankbadge{silver}{92.3} & \rankbadge{silver}{91.8} & \rankbadge{silver}{47.4} & \rankbadge{gold}{93.5} & \rankbadge{gold}{92.1} & \rankbadge{gold}{50.6} \\
& ViSA        & 78.1 & 72.6 & 14.8 & 82.1 & \rankbadge{silver}{95.4} & 28.3 & \rankbadge{bronze}{83.0} & 92.1 & \rankbadge{gold}{37.7} & \rankbadge{silver}{87.0} & \rankbadge{bronze}{95.6} & \rankbadge{bronze}{28.9} & \rankbadge{gold}{87.2} & \rankbadge{gold}{95.9} & \rankbadge{silver}{29.2} \\
& BTAD        & 83.3 & 72.6 & 18.5 & 89.1 & \rankbadge{gold}{93.3} & \rankbadge{silver}{49.7} & \rankbadge{silver}{91.6} & \rankbadge{bronze}{92.0} & \rankbadge{gold}{51.7} & \rankbadge{bronze}{91.2} & 91.9 & 16.6 & \rankbadge{gold}{91.8} & \rankbadge{silver}{92.2} & \rankbadge{bronze}{27.9} \\
& KSDD2       & \rankbadge{bronze}{93.5} & 94.9 & 69.2 & 92.1 & \rankbadge{silver}{99.1} & 70.7 & \rankbadge{silver}{95.9} & 96.1 & \rankbadge{bronze}{70.8} & 93.5 & \rankbadge{bronze}{97.9} & \rankbadge{silver}{71.2} & \rankbadge{gold}{96.2} & \rankbadge{gold}{99.3} & \rankbadge{gold}{72.6} \\
& MPDD        & 61.4 & 71.2 & 15.4 & \rankbadge{gold}{77.0} & \rankbadge{gold}{96.4} & \rankbadge{silver}{34.2} & \rankbadge{silver}{76.0} & \rankbadge{silver}{96.1} & \rankbadge{silver}{34.1} & 63.2 & 91.9 & \rankbadge{bronze}{31.4} & \rankbadge{bronze}{70.1} & \rankbadge{bronze}{94.1} & \rankbadge{gold}{34.3} \\
& MVTec-AD2   & 45.2 & 72.3 & 8.4 & \rankbadge{bronze}{51.3} & \rankbadge{silver}{85.4} & \rankbadge{bronze}{12.8} & 48.5 & 78.6 & 10.2 & \rankbadge{silver}{54.1} & \rankbadge{bronze}{83.2} & \rankbadge{silver}{14.6} & \rankbadge{gold}{59.4} & \rankbadge{gold}{89.5} & \rankbadge{gold}{17.5} \\
\cmidrule(lr){2-17}
\multirow{3}{*}{\textbf{Medical}} 
& Kvasir      & -- & 69.8 & 27.5 & -- & \rankbadge{bronze}{81.8} & \rankbadge{bronze}{53.8} & -- & 79.4 & 43.8 & -- & \rankbadge{silver}{85.4} & \rankbadge{silver}{63.9} & -- & \rankbadge{gold}{90.6} & \rankbadge{gold}{66.5} \\
& CVC-ColonDB & -- & 64.8 & 28.4 & -- & \rankbadge{bronze}{81.9} & \rankbadge{bronze}{41.4} & -- & 79.3 & 6.5 & -- & \rankbadge{silver}{82.1} & \rankbadge{silver}{42.1} & -- & \rankbadge{gold}{82.9} & \rankbadge{gold}{42.3} \\
& CVC-ClinicDB & -- & 51.2 & 24.4 & -- & \rankbadge{silver}{82.9} & \rankbadge{silver}{42.1} & -- & \rankbadge{bronze}{82.8} & \rankbadge{bronze}{40.9} & -- & 70.7 & 38.4 & -- & \rankbadge{gold}{90.7} & \rankbadge{gold}{57.2} \\
\bottomrule
\end{tabular}
}
\begin{tablenotes}
\small
\centering\item Note: For medical datasets, we do not have normal images available, so we are not reporting the I-AUC results.
\end{tablenotes}
\end{table*}

\subsection{Training Objective}
\label{Sec:training}
The training objective on the labeled auxiliary dataset $\mathcal{D}_a$ combines cross-modal alignment losses with routing regularization. For pixel-level supervision, we compute similarity between adapted patch tokens and text embeddings to generate anomaly maps $\mathbf{P} \in [0,1]^{H \times W}$, optimized against ground-truth masks $\mathbf{M}$ using focal and Dice losses that emphasize hard-to-classify patches and improve spatial consistency.

\begin{equation}
\mathcal{L}_{\text{seg}} = \mathcal{L}_{\text{focal}}(\mathbf{P}, \mathbf{M}) + \mathcal{L}_{\text{dice}}(\mathbf{P}, \mathbf{M})
\end{equation}

Without explicit regularization, routing weights converge to approximately equal values regardless of input, failing to learn meaningful pathway differentiation. We address this through routing regularization that encourages specialization:

\begin{equation}
\mathcal{L}_{\text{routing}} = \text{MSE}(w_n, 1-y) + \text{MSE}(w_a, y)
\end{equation}

where $y \in \{0,1\}$ is the binary anomaly label from auxiliary data, incentivizing normal samples to route through the normal adapter and vice versa. The complete objective is:

\begin{equation}
\mathcal{L} = \lambda_1 \mathcal{L}_{\text{seg}} + \lambda_2 \mathcal{L}_{\text{global}} + \lambda_3 \mathcal{L}_{\text{routing}}
\end{equation}

where $\mathcal{L}_{\text{global}}$ provides image-level supervision via cross-entropy. At inference (see Figure~\ref{fig:architecture}), detection relies solely on cross-modal similarity without routing regularization, maintaining efficiency equivalent to single-adapter methods.

\section{Experiments} \label{sec:experiment}
\noindent\textbf{Datasets.}
We evaluate on nine benchmarks spanning industrial and medical domains. Industrial datasets include MVTec-AD~\cite{bergmann2019mvtec} (15 categories), ViSA~\cite{zou2022spot} (12 categories), BTAD~\cite{mishra2021vt} (3 categories), KSDD2~\cite{Bozic2021COMIND} (surface defects), MPDD~\cite{jezek2021deep} (6 categories), and MVTec-AD2~\cite{heckler2025mvtec} (8 categories). 
In line with~\cite{zhou2024anomalyclip, qu2025bayesian}, we further evaluate on medical datasets: Kvasir~\cite{jha2019kvasir} (gastrointestinal polyps), CVC-ColonDB~\cite{tajbakhsh2015automated}, and CVC-ClinicDB~\cite{bernal2015wm}.
Following standard zero-shot protocols~\cite{zhou2024anomalyclip}, we train on ViSA and evaluate on all other datasets; for ViSA evaluation, we train on MVTec-AD. Medical results thus represent true cross-domain generalization from industrial to medical domain.\vspace{-0.2cm}\\

\noindent\textbf{Evaluation Metrics.}
We report three complementary metrics: Image-AUROC (I-AUC) for detection accuracy, Pixel-AUROC (P-AUC) for localization quality, and Pixel-F1 for precision-recall balance at optimal threshold. For medical datasets where normal reference images are unavailable, we report only pixel-level metrics. Results are averaged across categories within each dataset.\vspace{-0.2cm}\\

\noindent\textbf{Implementation Details.}
We employ frozen DINOv3-ViT-L/16~\cite{simeoni2025dinov3} and CLIP-ViT-L/14-336~\cite{radford2021learning} as visual and text encoders respectively. Each adapter branch uses bottleneck dimension 256 with LeakyReLU activations across four layers, yielding 28M trainable parameters. Training uses AdamW optimizer with learning rate $10^{-4}$ and batch size 64 for 20 epochs. Loss weights are set to $\lambda_1=0.5$, $\lambda_2=0.25$, $\lambda_3=0.1$ with routing temperature $\tau=0.1$. All images are resized to $512 \times 512$ for both training and inference. Model is trained on Nvidia-RTX 4090 with 24GB of RAM.\vspace{-0.2cm}\\

\noindent\textbf{Results.}
\label{sec:results}
Table~\ref{tab:main_results} reports our zero-shot anomaly detection performance against WinCLIP~\cite{jeong2023winclip}, AdaCLIP~\cite{cao2024adaclip}, AnomalyCLIP~\cite{zhou2024anomalyclip}, and Bayes-PFL~\cite{qu2025bayesian}. All baselines follow the same ZSAD training and evaluation protocol as ours (i.e. the same kind of auxiliary supervision, labels and masks). Our method ranks first on most metrics across both domains, highlighting the benefit of anomaly-aware vision-language adaptation coupled with DINOv3 features.

On \underline{industrial datasets}, our method obtains consistent improvements, achieving 93.5\% I-AUC and 92.1\% P-AUC on MVTec-AD with a substantial 50.6\% Pixel-F1, outperforming the second-best method by 1.2, 0.3, and 3.2 points respectively. The gains are more pronounced on challenging benchmarks: MVTec-AD2 shows 59.4\% I-AUC and 89.5\% P-AUC, representing improvements of 5.3 and 4.1 points over the nearest competitor. On KSDD2, we achieve 96.2\% I-AUC and 99.3\% P-AUC, demonstrating robust defect detection on surface inspection tasks. Notably, while AdaCLIP and AnomalyCLIP achieve competitive Pixel-F1 on BTAD (51.7\% and 49.7\%), our method maintains balanced performance across all three metrics rather than optimizing individual scores at the expense of others.

For \underline{medical imaging}, where normal reference images are unavailable and we report only pixel-level metrics, our method demonstrates strong cross-domain generalization. On CVC-ClinicDB, we achieve 90.7\% P-AUC and 57.2\% Pixel-F1, improving over the second-best by 7.8 and 15.1 points respectively. Kvasir results (90.6\% P-AUC, 66.5\% Pixel-F1) further confirm that our anomaly-aware adapters transfer to polyp segmentation without domain-specific fine-tuning.

\begin{figure}[t]
\centering
\includegraphics[width=\columnwidth]{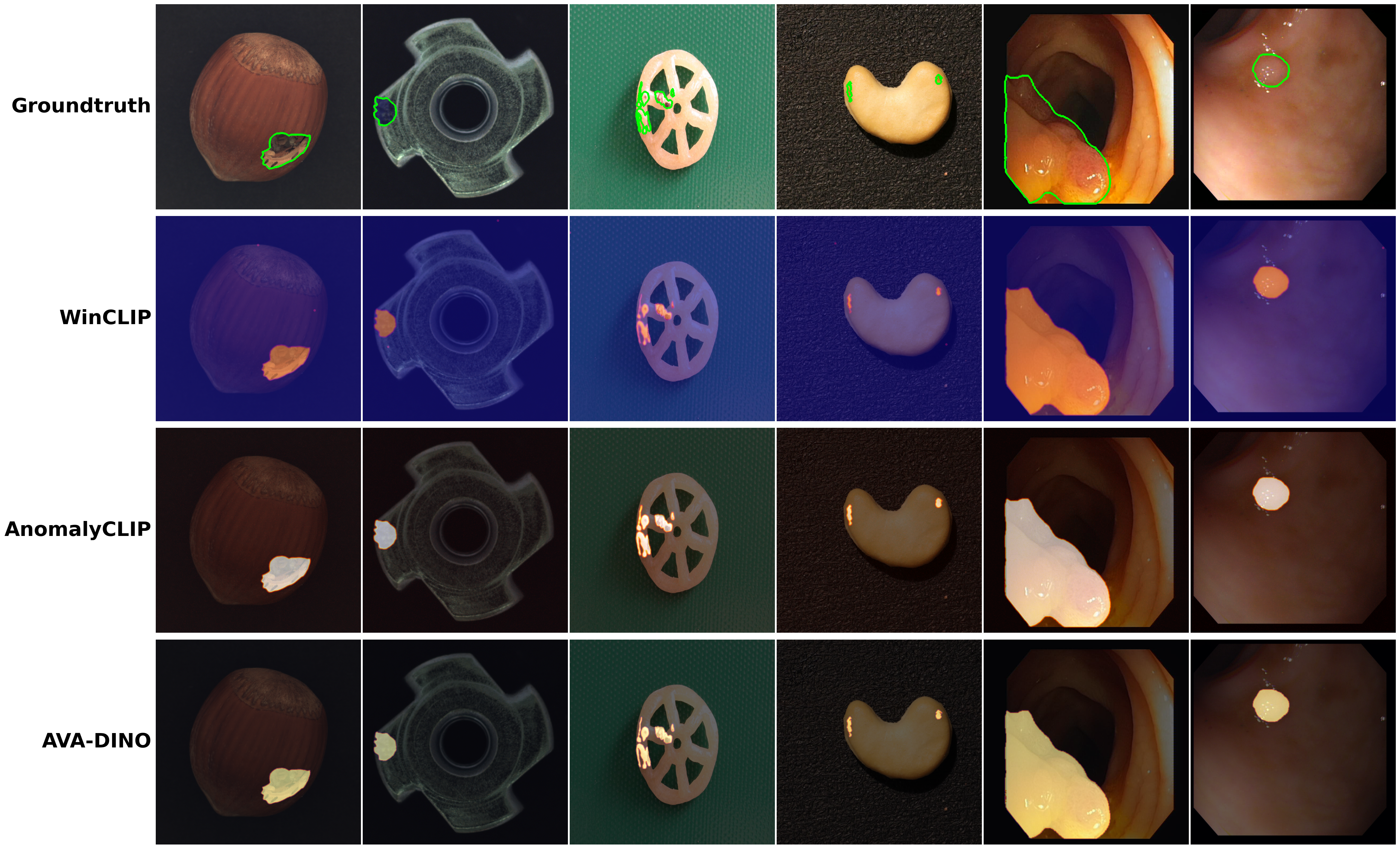}
\vspace{-0.5cm}
\caption{Anomaly localization results on industrial (columns 1-4) and medical (columns 5-6) samples. Ground truth boundaries shown in green.}
\label{fig:qualitative}
\end{figure}

Figure~\ref{fig:qualitative} provides \underline{qualitative} comparisons across industrial (hazelnut, metalnut, fryum, cashew from ViSA/MVTec) and medical (colonoscopy) samples. WinCLIP produces diffuse activations that fail to capture precise anomaly boundaries, while AnomalyCLIP tends toward over-segmentation on textured regions. Our method generates sharper, more accurate anomaly masks that better align with ground truth boundaries, particularly evident in the medical images where polyp delineation requires fine-grained spatial understanding.

\underline{Cross-Domain Analysis.} Our method demonstrates strong gains when transferring from industrial training data to medical benchmarks, 
with improvements of 7.8\% P-AUC on CVC-ClinicDB and 5.2\% on Kvasir over the second-best methods. Although industrial normals form narrow distributions (consistent shapes and colors), defects span diverse failure modes, mirroring medical imaging's high variance, explaining why our anomaly adapter, trained on spread defect patterns, transfers effectively to biomedical domains. In contrast, WinCLIP exhibits high variance across datasets (51.2\% to 94.9\% P-AUC), while our method maintains stable performance (82.9\% to 99.3\% P-AUC), indicating that text-guided routing provides robustness to domain shifts. The improvements across diverse anomaly types, from surface scratches in KSDD2 to polyps in colonoscopy images, validate that learning separate normal and anomaly pathways generalizes better than uniform adaptation approaches, establishing \methname as a general-purpose, domain-agnostic zero-shot anomaly detection framework.\\

\noindent\textbf{Ablation Study.}
We perform ablations on the MVTec-AD dataset to validate our design, isolating dual-branch adapters, text-guided routing, and routing regularization.

\underline{Component Analysis.} Table~\ref{tab:ablation_component} examines the impact of each proposed component. The single adapter baseline applies uniform transformation regardless of input characteristics. Introducing dual adapters with learnable routing (without text guidance) improves I-AUC by 0.8\%, demonstrating the benefit of separate pathways. Replacing learnable routing with text-guided routing improves performance, validating that CLIP's semantic understanding provides meaningful routing signals. However, without $\mathcal{L}_{\text{routing}}$, weights tend toward 50/50 regardless of input. Adding routing regularization enables proper pathway specialization, 
with our full model achieving 93.5\% I-AUC, 92.1\% P-AUC, and 50.6\% Pixel-F1 (+5.3\% over baseline).
\begin{table}[htbp]
\vspace{-0.3cm}
\centering
\caption{Component ablation on MVTec-AD. Arrows indicate improvement over baseline.}
\label{tab:ablation_component}
\resizebox{\columnwidth}{!}{%
\begin{tabular}{lccc}
\toprule
\textbf{Configuration} & \textbf{I-AUC} & \textbf{P-AUC} & \textbf{Pixel-F1} \\
\midrule
Single adapter (baseline) & 91.2 & 89.7 & 45.3 \\
Dual adapters + learnable routing & 92.0 \up{0.8} & 90.2 \up{0.5} & 46.5 \up{1.2} \\
Dual adapters + text-guided (w/o $\mathcal{L}_{\text{routing}}$) & 92.6 \up{1.4} & 90.8 \up{1.1} & 47.8 \up{2.5} \\
Full model & \textbf{93.5} \up{2.3} & \textbf{92.1} \up{2.4} & \textbf{50.6} \up{5.3} \\
\bottomrule
\end{tabular}%
}
\end{table}

\underline{Routing and Feature Analysis.} Table~\ref{tab:feature_layers} shows that multi-layer extraction from DINOv3 outperforms using only the last layer by 1.7\% P-AUC, confirming that anomaly detection benefits from both fine-grained and semantic representations. Table~\ref{tab:routing_weights} reports learned routing weights on test samples: normal samples receive $w_n=0.83$ while anomaly samples receive $w_a=0.79$, validating that $\mathcal{L}_{\text{routing}}$ enables meaningful specialization during training that generalizes at inference.

\begin{table}[htbp]
\centering
\small
\caption{Feature extraction strategy on MVTec-AD.}
\vspace{-0.3cm}
\label{tab:feature_layers}
\begin{tabular}{lcc}
\toprule
\textbf{Strategy} & \textbf{I-AUC} & \textbf{P-AUC} \\
\midrule
Last layer only & 91.8 & 90.4 \\
Multi-layer (Full) & \textbf{93.5} \up{1.7} & \textbf{92.1} \up{1.7} \\
\bottomrule
\end{tabular}
\vspace{0.2cm}
\centering
\caption{Routing weights on MVTec-AD test samples.}
\vspace{-0.3cm}
\label{tab:routing_weights}
\begin{tabular}{lcc}
\toprule
\textbf{Sample type} & \textbf{Avg $w_n$} & \textbf{Avg $w_a$} \\
\midrule
Normal & 0.83 & 0.17 \\
Anomaly & 0.21 & 0.79 \\
\bottomrule
\end{tabular}
\end{table}
\vspace{-1em}
\section{Conclusion}
\label{sec:conclusion}
We present \methname, a dual-branch adaptation framework for zero-shot anomaly detection that learns context-specific feature transformations. Unlike uniform adaptation, \methname uses separate normal and anomaly pathways, combined through text-guided routing with explicit regularization to enforce specialization. Experiments on industrial and medical benchmarks consistently outperform state-of-the-art methods, and routing weight analysis confirms meaningful specialization. Strong cross-domain generalization suggests that asymmetric adaptation is more effective than uniform approaches. Future work will extend \methname to few-shot and video anomaly detection.

\vfill\pagebreak



\bibliographystyle{IEEEbib}
\bibliography{strings,refs}

\begin{thebibliography}{10}

\bibitem{jeong2023winclip}
Jongheon Jeong, Yang Zou, Taewan Kim, Dongqing Zhang, Avinash Ravichandran, and Onkar Dabeer,
\newblock ``Winclip: Zero-/few-shot anomaly classification and segmentation,''
\newblock in {\em IEEE/CVF Conference on Computer Vision and Pattern Recognition (CVPR)}, 2023, pp. 19606--19616.

\bibitem{zhou2024anomalyclip}
Qihang Zhou, Guansong Pang, Yu~Tian, Shibo He, and Jiming Chen,
\newblock ``Anomalyclip: Object-agnostic prompt learning for zero-shot anomaly detection,''
\newblock in {\em International Conference on Learning Representations (ICLR)}, 2024, vol. 2024, pp. 49705--49737.

\bibitem{radford2021learning}
Alec Radford, Jong~Wook Kim, Chris Hallacy, Aditya Ramesh, Gabriel Goh, Sandhini Agarwal, Girish Sastry, Amanda Askell, Pamela Mishkin, Jack Clark, et~al.,
\newblock ``Learning transferable visual models from natural language supervision,''
\newblock in {\em International Conference on Machine Learning (ICML)}. PmLR, 2021, pp. 8748--8763.

\bibitem{simeoni2025dinov3}
Oriane Sim{\'e}oni, Huy~V Vo, Maximilian Seitzer, Federico Baldassarre, Maxime Oquab, Cijo Jose, Vasil Khalidov, Marc Szafraniec, Seungeun Yi, Micha{\"e}l Ramamonjisoa, et~al.,
\newblock ``Dinov3,''
\newblock {\em arXiv preprint arXiv:2508.10104}, 2025.

\bibitem{qu2025bayesian}
Zhen Qu, Xian Tao, Xinyi Gong, Shichen Qu, Qiyu Chen, Zhengtao Zhang, Xingang Wang, and Guiguang Ding,
\newblock ``Bayesian prompt flow learning for zero-shot anomaly detection,''
\newblock in {\em IEEE/CVF Conference on Computer Vision and Pattern Recognition (CVPR)}, 2025, pp. 30398--30408.

\bibitem{cao2024adaclip}
Yunkang Cao, Jiangning Zhang, Luca Frittoli, Yuqi Cheng, Weiming Shen, and Giacomo Boracchi,
\newblock ``Adaclip: Adapting clip with hybrid learnable prompts for zero-shot anomaly detection,''
\newblock in {\em European Conference on Computer Vision (ECCV)}. Springer, 2024, pp. 55--72.

\bibitem{aqeel2025contrastive}
Muhammad Aqeel, Danijel Sko{\v{c}}aj, Marco Cristani, and Francesco Setti,
\newblock ``A contrastive learning-guided confident meta-learning for zero shot anomaly detection,''
\newblock in {\em IEEE/CVF International Conference on Computer Vision (ICCV)}, 2025, pp. 1452--1461.

\bibitem{houlsby2019parameter}
Neil Houlsby, Andrei Giurgiu, Stanislaw Jastrzebski, Bruna Morrone, Quentin De~Laroussilhe, Andrea Gesmundo, Mona Attariyan, and Sylvain Gelly,
\newblock ``Parameter-efficient transfer learning for nlp,''
\newblock in {\em International Conference on Machine Learning (ICML)}. PMLR, 2019, pp. 2790--2799.

\bibitem{chen2022adaptformer}
Shoufa Chen, Chongjian Ge, Zhan Tong, Jiangliu Wang, Yibing Song, Jue Wang, and Ping Luo,
\newblock ``Adaptformer: Adapting vision transformers for scalable visual recognition,''
\newblock {\em Advances in Neural Information Processing Systems (NeurIPS)}, vol. 35, pp. 16664--16678, 2022.

\bibitem{bergmann2019mvtec}
Paul Bergmann, Michael Fauser, David Sattlegger, and Carsten Steger,
\newblock ``Mvtec ad--a comprehensive real-world dataset for unsupervised anomaly detection,''
\newblock in {\em IEEE/CVF Conference on Computer Vision and Pattern Recognition (CVPR)}, 2019, pp. 9592--9600.

\bibitem{zou2022spot}
Yang Zou, Jongheon Jeong, Latha Pemula, Dongqing Zhang, and Onkar Dabeer,
\newblock ``Spot-the-difference self-supervised pre-training for anomaly detection and segmentation,''
\newblock in {\em European Conference on Computer Vision (ECCV)}. Springer, 2022, pp. 392--408.

\bibitem{mishra2021vt}
Pankaj Mishra, Riccardo Verk, Daniele Fornasier, Claudio Piciarelli, and Gian~Luca Foresti,
\newblock ``Vt-adl: A vision transformer network for image anomaly detection and localization,''
\newblock in {\em IEEE 30th International Symposium on Industrial Electronics (ISIE)}. IEEE, 2021, pp. 01--06.

\bibitem{Bozic2021COMIND}
Jakob Bo{\v{z}}i{\v{c}}, Domen Tabernik, and Danijel Sko{\v{c}}aj,
\newblock ``{Mixed supervision for surface-defect detection: from weakly to fully supervised learning},''
\newblock {\em Computers in Industry}, 2021.

\bibitem{jezek2021deep}
Stepan Jezek, Martin Jonak, Radim Burget, Pavel Dvorak, and Milos Skotak,
\newblock ``Deep learning-based defect detection of metal parts: evaluating current methods in complex conditions,''
\newblock in {\em 13th International Congress on Ultra Modern Telecommunications and Control Systems and Workshops (ICUMT)}. IEEE, 2021, pp. 66--71.

\bibitem{heckler2025mvtec}
Lars Heckler-Kram, Jan-Hendrik Neudeck, Ulla Scheler, Rebecca K{\"o}nig, and Carsten Steger,
\newblock ``The mvtec ad 2 dataset: Advanced scenarios for unsupervised anomaly detection,''
\newblock {\em arXiv preprint arXiv:2503.21622}, 2025.

\bibitem{jha2019kvasir}
Debesh Jha, Pia~H Smedsrud, Michael~A Riegler, P{\aa}l Halvorsen, Thomas De~Lange, Dag Johansen, and H{\aa}vard~D Johansen,
\newblock ``Kvasir-seg: A segmented polyp dataset,''
\newblock in {\em International Conference on Multimedia Modeling (MMM)}. Springer, 2019, pp. 451--462.

\bibitem{tajbakhsh2015automated}
Nima Tajbakhsh, Suryakanth~R Gurudu, and Jianming Liang,
\newblock ``Automated polyp detection in colonoscopy videos using shape and context information,''
\newblock {\em IEEE Transactions on Medical Imaging}, vol. 35, no. 2, pp. 630--644, 2015.

\bibitem{bernal2015wm}
Jorge Bernal, F~Javier S{\'a}nchez, Gloria Fern{\'a}ndez-Esparrach, Debora Gil, Cristina Rodr{\'\i}guez, and Fernando Vilari{\~n}o,
\newblock ``Wm-dova maps for accurate polyp highlighting in colonoscopy: Validation vs. saliency maps from physicians,''
\newblock {\em Computerized Medical Imaging and Graphics (CMGI)}, vol. 43, pp. 99--111, 2015.

\end{thebibliography}

\end{document}